\documentclass[preprint,12pt]{elsarticle}

\usepackage{graphicx}
\usepackage{amssymb}

\usepackage{lineno}
\usepackage[utf8]{inputenc}
\usepackage[english]{babel}
\usepackage{csquotes}
\usepackage{amsmath}
\usepackage{color}
\usepackage{graphicx}
\usepackage{setspace}
\usepackage{amsthm}
\usepackage{hyperref}
\usepackage{outlines}
\usepackage{caption}
\usepackage{amsfonts}
\usepackage{amssymb}
\usepackage{booktabs}
\usepackage{float}
\usepackage{tikz}
\usepackage{graphicx}
\usepackage{subfigure}
\usepackage{bm}
\usepackage[normalem]{ulem}
\usepackage[nottoc,notlot,notlof]{tocbibind}
\usepackage{parskip} 
\usepackage{multirow} 

\usepackage{newfloat}

\DeclareFloatingEnvironment[name={List}]{enumcnt}

\journal{ } 

\begin{document}

\begin{frontmatter}


\title{Data Assimilation in the Latent Space of a Neural Network}



\author[a]{Maddalena Amendola}
\author[a]{Rossella Arcucci\corref{cor1}}  
    \ead{r.arcucci@imperial.ac.uk} 
    \cortext[cor1]{Corresponding author}
\author[b]{Laetitia Mottet}
\author[a]{C\'esar Quilodr\'an Casas}
\author[c]{Shiwei Fan}
\author[a,b]{Christopher Pain}
\author[d]{Paul Linden}
\author[a]{Yi-Ke Guo}

\address[a]{Data Science Institute, Department of Computing, Imperial College London, UK}
\address[c]{Department of Chemistry, University of Cambridge, UK}
\address[b]{Department of Earth Science $\&$ Engineering, Imperial College London, UK}
\address[d]{Department of Applied Mathematics and Theoretical Physics, University of Cambridge, UK}

\begin{abstract}
There is an urgent need to build models to tackle Indoor Air Quality issue. Since the model should be accurate and fast, Reduced Order Modelling technique is used to reduce the dimensionality of the problem. The accuracy of the model, that represent a dynamic system, is improved integrating real data coming from sensors using Data Assimilation techniques. In this paper, we formulate a new methodology called Latent Assimilation that combines Data Assimilation and Machine Learning. We use a Convolutional neural network to reduce the dimensionality of the problem, a Long-Short-Term-Memory to build a surrogate model of the dynamic system and an Optimal Interpolated Kalman Filter to incorporate real data. Experimental results are provided for CO\textsubscript{2} concentration within an indoor space. This methodology can be used for example to predict in real-time the load of virus, such as the SARS-COV-2, in the air by linking it to the concentration of CO\textsubscript{2}.
\end{abstract}


\end{frontmatter}


\section{Introduction}
Urbanisation is the process for which people move from rural zone to urban zone changing their habits. This process grows year by year: about half of the global population already lives in urban areas and by 2050 two-thirds of the world's people are expected to live in urban areas. Urbanisation process has led to an increase in building, human activities and energy consumption causing environmental degradation. High building densities and the low presence of vegetation impair the air quality and circulation. People who live in such areas are hesitant to open the windows of their house thinking that this can led to an increment of pollution in their habitation. A solution from their point of view is to use air-conditioning increasing in this way the energy consumption. This is a vicious cycle of increased urban emissions of heat, pollutants and greenhouse gases and an associated increase in energy demand.

The scope of the MAGIC\footnote{http://www.magic-air.uk/home.html} (Managing Air for Green Inner Cities) project is to study and build systems to assist reduction in energy demand through natural ventilation~\cite{song2018natural}. To this aim, there is the need to use systems with high accuracy in predicting air flows and air pollution concentration. These systems use the Large Eddy Simulation method within the Computational Fluids Dynamics (CFD) software: Fluidity~\cite{AMCG2015}. Fluidity is an open source, general purpose, multi-phase computational fluid dynamics code capable of numerically solving the Navier-Stokes equations and advection-diffusion equations on arbitrary unstructured finite-element meshes. Fluidity is used in a number of different scientific areas including geophysical fluid dynamics, ocean modelling, mantle convection and air pollution.

Numerical simulation has been widely applied in many fields including environmental sciences, aerospace engineering, bio-medicine and industrial design. It provides powerful technical support for solving industrial problems and making scientific research in these fields. However, high fidelity numerical simulations of complex systems consume vast time and computing resources. When real data collected by instruments (i.e. sensors) are available, it is possible to use them to improve the accuracy of the prediction. The integration is made up by Data Assimilation techniques. 

Data Assimilation (DA) is an approach for fusing data (observations) with prior knowledge (e.g., mathematical representations of physical laws; model output) to obtain an estimate of the distribution of the true state of a process~\cite{wikle2007bayesian}. In order to perform DA, one needs observations (i.e., a data or measurement model), a background (i.e., a priori state or process model) and information about the distribution of the errors on these two. For those applications, where the background is defined in big computational grids which lead to a big data problem sometimes impossible to handle without introducing approximations or space reductions, Reduced Order Modelling (ROM) techniques are used~\cite{arcucci2019optimal, arcucci2019domain}. 

ROM allows to speed up the dynamic model and the DA process. Popular approaches to reduce the domain are the Principal Component Analysis (PCA) and the Empirical Orthogonal Functions (EOF) technique both based on a Truncated Singular Value Decomposition (TSVD) analysis~\cite{hansen2006deblurring}. The simplicity and the analytic derivation of those approaches are the main reasons behind their popularity in atmospheric and ocean science. However, despite those powerful approaches, the accuracy of the obtained solution exhibits a severe sensibility to the variation of the value of the truncation parameters. This issue introduces a severe drawback to the reliability of these approaches, hence their usability in operative software in different scenarios~\cite{hannachi2004primer}.

An approach to reduce the dimensionality maintaining information of the data is the Neural Network (NN), precisely the AutoEncoders~\cite{mack2020attention, wu2020data}. NNs have the ability to fit functions and they can fit almost any unknown function theoretically. That is the ability which makes it possible for neural networks to face complex problems. AutoEncoders with non-linear encoder functions and non-linear decoder functions can thus learn a more powerful non-linear generalisation of methods based on TSVD. In the latent space, the evolution of the transformed state variables defined in time, can be learned using Recurrent Neural Networks (RNN)~\cite{casas2020reduced, arcucci2020neural}. In the present work, we propose a new methodology which we called Latent Assimilation (LA). It consists in reducing the dimensionality with NN and perform both prediction through a surrogate dynamic model and DA directly in the latent space. In the latent space, the surrogate dynamic system is built by a RNN.

\section{Related Work and contribution of the present work}
The future challenges of Numerical Weather Prediction (NWP) is to include more accurate initial conditions that take advantage of the increasing volume of real-time observations, and improve the post-processing of model outputs, amongst others~\cite{boukabara2019leveraging}. To answer this need, Neural network (NN) for correction of error in forecasting have been extensively studied~\cite{babovic2000global,babovic2001neural,babovic2002data}. However, the error correction by NN does not have a direct relation with the updated model system at each step and the training is not on the results of the assimilation process. 

A framework for integration of NN with physical models by Data Assimilation (DA) algorithms is described in~\cite{zhu2019model}: the NNs are iteratively trained when observed data are updated. Unfortunately, this approach presents a limit due to the time complexity of the numerical models involved, which limits the use of the forecast model for large data problems. An approach for employing artificial neural networks (NNs) to emulate the Local Ensemble Transform Kalman Filter (LETKF) as a method of data assimilation is presented in~\cite{cintra2018data}. Deep learning and Data Assimilation technologies are also combined to predict the production of gas from mature gas wells in~\cite{loh2018deep}. The authors used a modified deep Long Short-Term Memory (LSTM) model as their prediction model in the Ensemble KF framework for parameter estimation. A Neural Network is integrated into a conventional DA in~\cite{zhu2019model}: deep learning shows great advantage in function approximations which have unknown model and strong non-linearity. The authors used NNs to characterise the structural model uncertainty. The NN is implemented in an End-to-End (E2E) approach and its parameters are iteratively updated with coming observations by applying the DA method.

A framework which performs fast data assimilation with sufficient accuracy for open ocean is proposed in~\cite{quilodranfast}. Speed improvement is achieved by performing the data assimilation on a reduced-space rather than on a full-space. A dimension reduction of the full-state is made by an Empirical Orthogonal Function (EOF) analysis while retaining most of the explained variance. Analysis of EOFs can be used to identify structures in geophysical data which hold a large part of the variance. In this framework, the assimilation is performed in the control space. EOFs analysis has become a fundamental tool in atmosphere, ocean, and climate science for data diagnostics and dynamical mode reduction. Each of these applications exploits the fact that EOFs allow a decomposition of a data function into a set of orthogonal functions, which are designed so that only a few of these functions are needed in lower-dimensional approximations. Furthermore, since EOFs are the eigenvectors of the error co-variance matrix, its condition number is reduced as well. Nevertheless, the accuracy of the solution obtained by truncating EOFs exhibits a severe sensibility to the variation of the value of the truncation parameter, so that a suitably choice of the number of EOFs is strongly recommended. This issue introduces a severe drawback to the reliability of EOFs truncation, hence to the usability of the operative software in different scenarios. A powerful solution to this is to use a Tikhonov regularisation which reveals to be more appropriate than truncation of EOFs~\cite{arcucci2019optimal}. 

Neural networks have tremendous ability to fit functions and they can fit almost any unknown function theoretically. That is the ability which makes it possible for neural networks to model complex flows. The complex computations involving matrices is reduced by factorising the representation deriving a latent state used from the Kalman Filter in~\cite{becker2019recurrent}. The authors also used a linear dynamic model to compute, i.e predict, the next timestep. A variational AutoEncoder capable to generate trajectories from a latent space where the dynamics is linear is presented in~\cite{watter2015embed}.

In this paper, we propose a new methodology that use the NNs to reduce the space and perform the assimilation of the sensors data in the latent space. Specifically, we use a Convolutional AutoEncoder to reduce the domain and we perform an Optimal Interpolated Kalman Filter in the latent space.

In this paper, we make the following contributions:
\begin{itemize}
    \item We have designed a novel data assimilation technology, we called Latent Assimilation (LA), mainly composed by an AutoEncoder, a surrogate model and an Optimal Kalman Filter. The Latent Assimilation model performs the prediction of the flows and the assimilation of observed data through a Kalman Filter in the latent space.
    \item We have developed a Convolutional AutoEncoder to reduce the space where the surrogate model will work and where we perform the assimilation of the observation using the Optimal Interpolated Kalman Filter. We have chosen to use an encoder-decoder model instead of Principal Component Analysis (PCA) since neural networks maintain non-linearities and perform better in modelling flows;
    \item We have built a Recurrent Neural Network (LSTM) to emulate a Computational Fluid Dynamics (CFD) simulation in the latent space of an AutoEncoder: the trained LSTM represents the surrogate model to predict the CO\textsubscript{2} concentration in a room;
    \item We prove that our novel Latent Assimilation model answers the needs of accuracy, stability and efficiency required by real-time applications.
    \item We have developed a software written in python to test the Latent Assimilation model. The LA code and the pre-processed data can be downloaded using the link:
    
    \url{https://github.com/DL-WG/LatentAssimilation}.
\end{itemize}

Experimental results are provided for pollutant dispersion within an indoor space. This methodology can be used for example to predict in real-time the load of virus, such as the SARS-COV-2, in indoor spaces by linking it to the concentration of CO\textsubscript{2}~\cite{peng2020exhaled}.
\section{Data Assimilation}\label{Sec:KF}
In this section, we introduce  the concept of Data Assimilation (DA) and the Kalman Filter (KF) which is one of the most used approach for DA.

DA merges the estimated state $x_{t} \in \mathbb{R}^{n}$ of a discrete-time dynamic process at time $t$: 
\begin{equation}\label{eq:statevector}
    x_{t+1} = M_{t+1}x_{t} + w_{t}
\end{equation}
with an observation $y_{t} \in \mathbb{R}^{m}$:
\begin{equation}\label{eq:observation}
    y_{t} = H_{t}x_{t} + v_{t}
\end{equation}
where $M_{t+1}$ is a dynamic linear operator and $H_{t}$ is the observation operator. The vectors $w_{t}$ and $v_{t}$ represent the process and observation errors, respectively. They are usually assumed to be independent, white-noise processes with Gaussian probability distributions:
\begin{center}
    $w_t \sim \mathcal{N}(0, Q_t)$,\quad  $v_t \sim \mathcal{N}(0, R_t)$
\end{center}
where $Q_t$ and $R_t$ are called errors covariance matrices of the model and the observations, respectively.

DA tries to answer questions such as "what can be said about the value of an unknown variable $x_t$ that represents the evolution of a system, if we have some measured data $y_t$ and a model $M$ of the underlying mechanism that generated the data?". This is the Bayesian context, where we seek a quantification of the uncertainty in our knowledge of the parameters that, according to Bayes’ rule takes the form
\begin{equation}
  p\left( x_t|y_t \right)  = \frac{p\left( y_t|x_t \right) p\left( x_t\right)}{p\left( y_t \right)}
\end{equation}
Here, the physical model is represented by the conditional probability (also known as the likelihood) $p\left(y_t | x_t\right)$, and the prior knowledge of the system by the term $p\left(x_t\right)$. The denominator is considered as a normalising factor and represents the total probability of $y_t$. 
DA is a Bayesian inference that combines the state $x_t$ with $y_t$ at each given time. The Bayes theorem conducts to the estimation of $x_t^{a}$ which maximise a probability density function given the observation $y_t$ and a prior from $x_t$. This approach is implemented in one of the most popular DA methods which is the Kalman Filter (KF)~\cite{kalman1960new} which mainly consists of two steps: a prediction (equation~\eqref{Eq: StartKalman}) and a correction (equations~\eqref{eq:kalmangain}-\eqref{eq:kalmananalysis}) steps. 
The goal of the KF is to compute an optimal \textit{a posteriori} estimate, $x^a_t$, which is a linear combination of an \textit{a priori} estimate, $x_t$, and a weighted difference between the actual measurement, $y_t$, and the measurement prediction, $H_tx_t$ as described in equation~\eqref{eq:kalmananalysis}. 

\begin{enumerate}
    \item \textbf{Prediction}:
        \begin{equation}
            x_{t+1} = M_{t+1}x^a_t
            \label{Eq: StartKalman}
        \end{equation}
    \item \textbf{Correction}:
        \begin{equation}\label{eq:kalmangain}
            K_{t+1} = Q_{t+1}H_{t+1}^T(H_{t+1}Q_{t+1}H_{t+1}^T + R_{t+1})^{-1}
        \end{equation}
        \begin{equation}\label{eq:kalmananalysis}
            x^a_{t+1} = x_{t+1} + K_{t+1}(y_{t+1} - Hx_{t+1})
        \end{equation}
\end{enumerate}
For big data problems, KF is usually implemented in a simplified version as an Optimal Interpolation method~\cite{asch2016data} for which the covariance matrix $Q_t=Q$ is fixed at each timestep $t$. \\
The prediction-correction cycle is complex and time-consuming and it mandates the introduction of simplifications, approximations or data reductions techniques. In the next section, we present the Latent Assimilation approach which consists in performing KF in the latent space of an Autoencoder with nonlinear encoder and nonlinear decoder functions. In the latent space, the dynamic system in equation~\eqref{Eq: StartKalman} is replaced by a surrogate model built with a RNN.

\section{Latent Assimilation}\label{Sec:LA}
\begin{figure}[h!]
    \centering
    \includegraphics[width=1\textwidth]{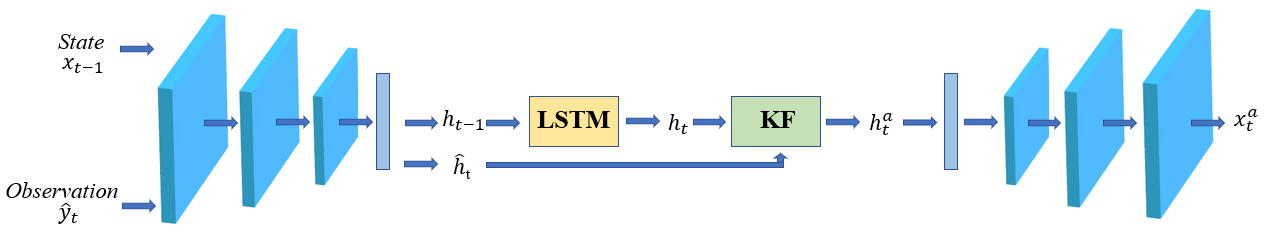}
    \caption{Latent Assimilation model workflow. Let assume that we want to predict the state of the system at time $t$ and we assume that the LSTM needs one observation back to predict the next timestep. The input of the system is the state $x_{t-1}$. We encode $x_{t-1}$ producing its encoded version $h_{t-1}$. From $h_{t-1}$ we compute $h_t$ through LSTM. To perform the Kalman Filter, we need the observation $\hat{y}_t$ at timestep $t$. We encode $y_t$ and we combine the result, $\hat{h}_t$, with the prediction ${h}_t$ through the KF. The result $h^a_t$ is the updated prediction. We report the updated prediction in its physical space through the Decoder, producing $x^a_t$.} \label{Fig: Model}
\end{figure}

Latent Assimilation is a model that implements the idea of assimilating real data in the Latent Space of a Neural Network (NN).
Instead of using PCA or others mathematical approaches to reduce the space, we model the reduction with non-linear transformations using Deep NNs. Specifically, we choose to use Convolutional Autoencoder to reduce the space.
The model is divided into four main parts: 
\begin{enumerate}
    \item Dimensionality reduction: the physical space is transformed in a latent space of smaller dimension by a Convolutional Autoencoder;
    \item Surrogate model: a surrogate of the CFD is built in the latent space by a Recurrent Neural Network;
    \item Data Assimilation: observed data are assimilate in the surrogate of the CFD by a Kalman Filter;
    \item Physical space: the results of the DA in the latent space are then reported in the physical space through a Decoder.
\end{enumerate}
\noindent Figure \ref{Fig: Model} shows the work flows of the Latent Assimilation model.

\subsection{Dimensionality reduction} 
The dimensionality reduction is implemented by an AutoEncoder (AE). AEs are usually used for dimensionality reduction or feature learning.
To use the autoencoder for dimensionality reduction, the encoder function must returns an output with lower dimension with respect to the input. This kind of autoencoder are called \textit{undercomplete}. Learning an undercomplete representation forces the autoencoder to capture the most salient features of the training data.
One type of undercomplete autoencoder is the Convolutional autoencoder. As we can deduce from the name, this autoencoder uses the Convolutional operation. Thanks to the convolutional operation, the network takes into account the spatial information: they are specially used with images or grid data. Usually, the Convolutional AutoEncoders are composed by more than one convolutional layers, each followed by pooling layer to reduce the input \cite{goodfellow2016deep}.
Latent Assimilation implements a Convolutional Autoencoder 
which produces a representation of the state vector $x_t \in \mathbb{R}^{n}$ in \eqref{eq:statevector} in a ``latent'' state vector $h_t \in \mathbb{R}^{p}$ defined in a Latent Space where $p<n$. We denote with $f: \mathbb{R}^{n} \to \mathbb{R}^p $ the Encoder function
\begin{equation}
    h_t = f(x_t)
\end{equation}
which transforms the state $x_t$ in a latent variable $h_t$.

\begin{figure}[h!]
\centering
\includegraphics[width=0.6\textwidth]{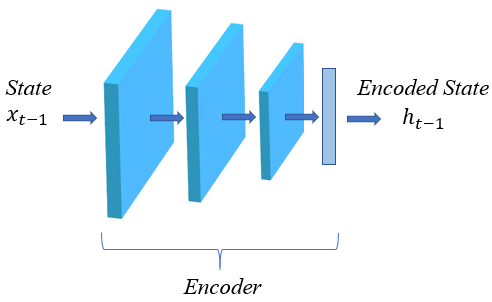}
\caption[Latent Assimilation Encoder]{Latent Assimilation - Encoder: $x_{t-1}$ is the input state and $h_{t-1}$ is the corresponding encoded state at timestep $t-1$.}
\label{Fig: encoder_LA}
\end{figure}

\subsection{Surrogate model}
\begin{figure}[h!]
\centering
\includegraphics[width=0.6\textwidth]{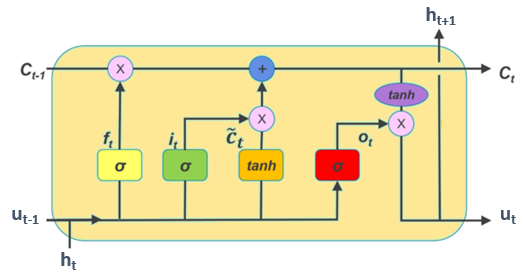}
\caption[Latent Assimilation LSTM]{Latent Assimilation - LSTM.}
\label{Fig: lstm_LA}
\end{figure}

In the latent space we perform a regression through a Long Short Term Memory (LSTM) function $l: \mathbb{R}^{p\times q} \to \mathbb{R}^p $
\begin{equation}\label{eq:surrogateModel}
    h_{t+1} = l(\bm{h_{t,q}})
\end{equation}
where $\bm{h_{t,q}}=\left\{ h_i \right\}_{i=t, \dots, t-q}$ is a sequence of $q$ encoded timesteps up to time $t$. 
The LSTM is a Recurrent Neural Network (RNN) with good performance with time-series data \cite{hochreiter1997long}. It is composed by gates and cells as shown if Figure~\ref{Fig: lstm_LA}. The gates decide which information should pass using a sigmoid function.
The LSTM is composed by four elements described below. In all formulas, $b$, $U$ and $W$ denote respectively the biases, input weights and recurrent weights for corresponding gate.
\begin{enumerate}
    \item \textbf{Forget gate}: it decides what information should pass via a logistic sigmoid $\sigma(x) = (1/(1+e^{-x}))$ unit
    \begin{equation}
        d_i^{(t)} = \sigma(b_i^d + \sum_j U_{i,j}^dh_j^{(t)} + \sum_j W_{i,j}^du_j^{(t-1)})
    \end{equation}
    where $h^{(t)}$ is the input and $u^{(t)}$ is the hidden layer. 
    
    \item \textbf{Input Gate}: it is similar to the Forget Gate but with its own parameters
    \begin{equation}
        c^{(t)}_i = \sigma(b^c_i + \sum_j U^c_{i,j}h^{(t)}_j + \sum_j W^c_{i,j}u^{(t-1)}_j)
    \end{equation}
    
    \item \textbf{Cell State}: the cell state is then updated using sigmoid and hyperbolic tangent functions
    \begin{equation}
        s_i^{i} = d^{(t)}_i s^{(t-1)}_i + c^{(t)}_i \sigma (b_i + \sum_jU_{i,j}h^{(t)}_j + \sum_jW_{i,j} u^{(t-1)}_j)
    \end{equation}
    
    \item \textbf{Output Gate}: it decides what the value of the next hidden state using a sigmoid function on the previous hidden state and the current input and an hyperbolic tangent function on the newly modified cell state. 
    \begin{equation}
        q^{(t)}_i = \sigma(b^o_i + \sum_j U^o_{i,j} h^{(t)}_j + \sum_j W^o_{i,j} u^{(t-1)}_j)
    \end{equation}
    \begin{equation}
        h^{(t+1)}_i = \tanh(s^{(t)}_i)q^{(t)}_i
    \end{equation}
\end{enumerate}

\subsection{Data Assimilation} 
\begin{figure}[h!]
\centering
\includegraphics[width=0.9\textwidth]{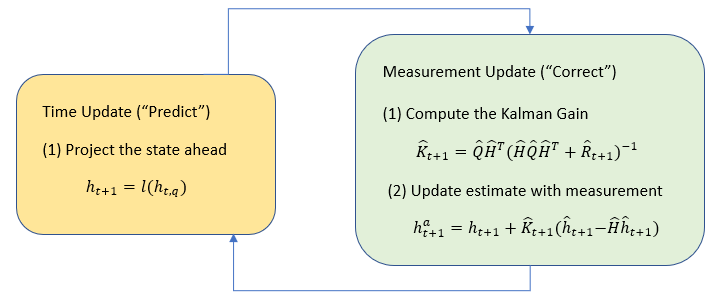}
\caption[Latent Assimilation Kalman Filter loop]{Latent Assimilation - Kalman Filter loop: the yellow block is the prediction phase (LSTM) and the green block is the correction phase (KF).}
\label{Fig: KF_LA}
\end{figure}

The assimilation is performed in the latent space. In order to merge the observations in \eqref{eq:observation} with the ``latent'' state vector $h_t$, the observations are processed by the Encoder in the same way as the state vector.
As $y_t\in \mathbb{R}^m$ where usually $m \leq n$, i.e. the observations are usually held or measured in just few point in space, the observations vector $y_t$ is interpolated in the state space $\mathbb{R}^n$ obtaining $\hat{y}_t \in \mathbb{R}^n$. The observations $\hat{y}_t$ are then processed in the same way as the state vector trough $f$:
\begin{equation}
    \hat{h}_t = f(\hat{y}_t)
\end{equation}
The ``latent'' observations $\hat{h}_t$, transformed by the Encoder in the latent space, are then assimilated by the prediction-correction steps as described in equations \eqref{Eq: LSTM_KF}-\eqref{Eq:LAfinal} ad as shown in Figure \ref{Fig: KF_LA}:
\begin{enumerate}
    \item \textbf{Prediction}:
        \begin{equation}
            h_{t+1} = l(\bm{h_{t, q}})
            \label{Eq: LSTM_KF}
        \end{equation}
    \item \textbf{Correction}:
        \begin{equation}
            \hat{K}_{t+1} = \hat{Q}\hat{H}^T(\hat{H}\hat{Q}\hat{H}^T + \hat{R}_{t+1})^{-1}
        \end{equation}
        \begin{equation}\label{Eq:LAfinal}
            h^a_{t+1} = h_{t+1} + \hat{K}_{t+1}(\hat{h}_{t+1} - \hat{H}h_{t+1})
        \end{equation}
\end{enumerate}

\noindent where $l$ in \eqref{Eq: LSTM_KF} is the surrogate model defined in \eqref{eq:surrogateModel} computed by the LSTM,
$\hat{Q}$ and $\hat{R}$ are the errors covariance matrices of the transformed background $h_t$ and observations $\hat{h}_t$ respectively: they are computed directly in the latent space. 
The background covariance matrix $\hat{Q}$ is computed with a sample of $s$ model state forecasts $\bm{h}$ that we set aside as background such that:
\begin{equation}\label{eq:covarianProcedure}
    \bm{h} = [h_1, ..., h_s]  \in \mathbb{R}^{p \times s}, \quad  V = (\bm{h} - \bar{h} ) \in \mathbb{R}^{p \times s}
\end{equation}
where $  \bar{h} $ is the mean of the sample of background states, then $ \hat{Q}= VV^T$. The observations errors covariance matrix $\hat{R}$ can be computed with the same process than in equation~\eqref{eq:covarianProcedure} by replacing $h_t$ with $\hat{h}_t$ $\forall t$

\begin{equation}\label{eq:covarianceRProcedure}
    \bm{\hat{h}} = [\hat{h}_1, ..., \hat{h}_s]  \in \mathbb{R}^{p \times s}, \quad  \hat{V} = (\bm{\hat{h}} - \bar{\hat{h}} ) \in \mathbb{R}^{p \times s}
\end{equation}
where $  \bar{\hat{h}} $ is the mean of the sample of observations, then $ \hat{R}= \hat{V}\hat{V}^T$.
The covariance matrix $\hat{R}$ can be estimated by evaluations of measurements (instruments) errors 
\begin{equation}
    \hat{R}= \sigma I \in \mathbb{R}^{p \times p}
\end{equation}
where $0<\sigma <1$ and $I \in \mathbb{R}^{p \times p}$ denotes the identity matrix \cite{asch2016data}.
$\hat{K}$ is the Kalman Gain matrix defined in the latent space and $\hat{H}$ is the observation operator.

\subsection{Physical space}
The results of the DA in the latent space are then reported in the physical space through the Decoder, applying the function $g: \mathbb{R}^p \to \mathbb{R}^{n}$ to compute
\begin{equation}
    x^a_{t+1} = g(h^a_{t+1}).
\end{equation}
The Decoder is almost a mirror of the Encoder: it is composed of a Fully Connected Layer followed by some Convolutional Layers.

\begin{figure}[h!]
\centering
\includegraphics[width=0.6\textwidth]{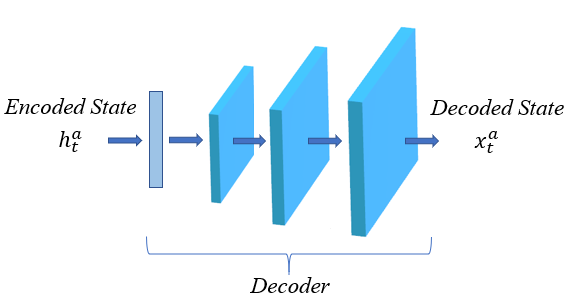}
\caption[Latent Assimilation Decoder]{Latent Assimilation - Decoder: $h^a_t$ is the encoded updated state and $x^a_t$ is the decoded updated state.}
\label{Fig: decoder_LA}
\end{figure}

\subsection{Latent Assimilation code}

The code is written in Python and is available at the following link: \url{https://github.com/DL-WG/LatentAssimilation}.
The LatentAssimilation folder is composed by different subfolders:
\begin{itemize}
    \item DataSet: it contains the Structured dataset divided in train and test;
    \item PreProcess: it contains all the code written to extract the Structured dataset starting from the unstructured meshes. We used the python libraries math, numpy, vtktools and pyvista.
    \item AutoEncoder and LSTM: both folders contain the code used to find the structure of the model and the hyper-parameters for the Structured dataset. All results are stored and also visualized in AnalysisLS7 jupyter notebook. We used python libraries such as numpy, sklearn, pandas and tensorflow.
    \item Data Assimilation: here there is all the observation data preprocessing, the Kalman Filter and the LatentAssimilation module which performs the assimilation in the Latent Space and it prints the table of the results.
\end{itemize}

\noindent In the next Section we apply Latent Assimilation to the problem of assimilating data to improve the prediction of air flows and indoor pollution transport in a real scenario \cite{song2018natural}. We show the performance of the model step by step and we compare results with a standard DA performed in the physical space.

\section{Set up of a real test case}
\subsection{CFD simulation}
\begin{figure}
    \centering
    \includegraphics[width=0.8\textwidth]{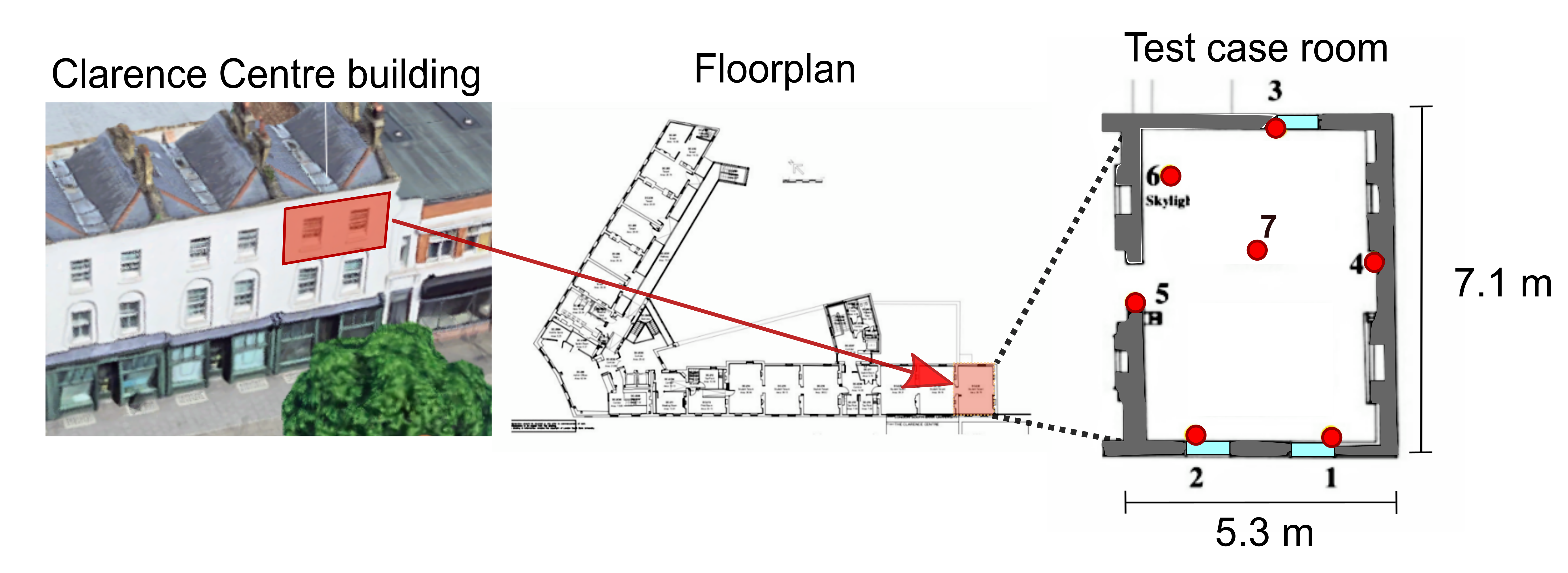}
    \caption{Test case room located in Clarence Centre building, London, UK. Red dots denote the location of the 7 sensors used during the field experiment. Blue rectangles show the location of the three windows~\cite{song2018natural}.}
    \label{Fig:ClarenceCentre}
\end{figure}

The LA model presented in Section~\ref{Sec:LA} is applied to real data collected in the context of the MAGIC project~\cite{song2018natural}: external and internal air quality measurements were performed within a naturally ventilated office room located at the top floor of the three-storey Clarence Centre building, Borough of Southwark, London, UK (Figure~\ref{Fig:ClarenceCentre}). The room has two windows facing a busy road (London Road), one window facing a traffic-free courtyard and a skylight in the ceiling. Seven sensors located in different positions were used to record, amongst others, the indoor temperature and CO\textsubscript{2} concentration, with a sampling rate of 1 minute. The three windows were opened during 25 minutes to look at cross ventilation effect on the decay of temperature and CO\textsubscript{2} concentration. During the whole period of the experiment, the predominant wind was a south-westerly wind.

\begin{figure}
    \centering
    \includegraphics[width=0.8\textwidth]{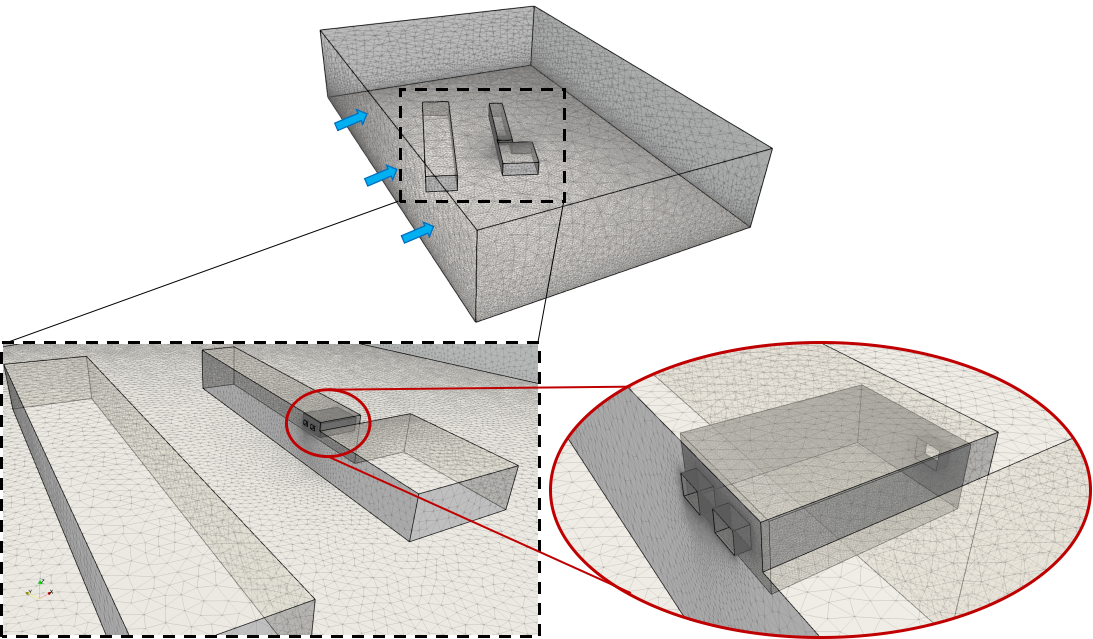}
    \caption{Computational domain and surface mesh of the area of interest showing the Clarence Centre and the upwind building as well as the test case room. The blue arrows denote the wind direction.}
    \label{Fig:ClarenceCentre_mesh}
\end{figure}

To replicate the field study experiment, a numerical simulation has been performed using the Computational Fluid Dynamics (CFD) software Fluidity (\url{http://fluidityproject.github.io/}). The same CFD simulation has been used in a previous paper~\cite{Tajnafoi2021} and only the main details of the CFD setup are re-called here. The computational domain includes the Clarence Centre building as well as the immediate upwind building, and the test room office as shown in Figure~\ref{Fig:ClarenceCentre_mesh} in order to replicate the cross ventilation scenario done during the field study. The mesh generated is an unstructured tetrahedral mesh composed by 400,000 nodes (Figure~\ref{Fig:ClarenceCentre_mesh}). The initial and boundary conditions are set to replicate the experimental conditions and are derived from the indoor sensors and the weather station used during the field study. The initial indoor CO\textsubscript{2} concentration is set equal to 1420 ppm, while the outdoor background CO\textsubscript{2} level is set equal to 400 ppm. The initial indoor and outdoor temperatures are equal to 19.5 $^{o}C$ and 9.1 $^{o}C$, respectively. The inlet velocity is following a log-law profile reaching 2.58 m/s at 28.5 m height. The simulation was run in parallel on 20 CPU and 15 minutes were simulated rendering approximately 3,500 timesteps. In this paper, the working variable of interest is the CO\textsubscript{2} concentration. It is worth noting that after the timestep 2,500 the concentration of CO\textsubscript{2} is low everywhere in the room since the room is completely ventilated.

\subsection{CFD data pre-processing}
\begin{figure}
    \centering
    \includegraphics[width=0.4\textwidth]{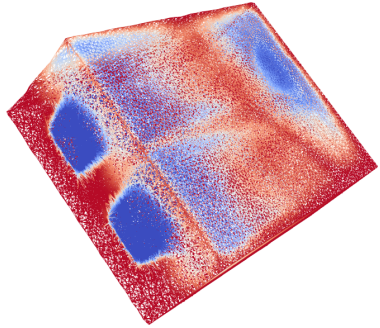}
    \caption{Unstructured mesh of the room coloured by CO\textsubscript{2} concentration. Blue colour means low CO\textsubscript{2} concentration, i.e. 400 ppm, and red colour indicates high value of concentration, i.e. 1420 ppm.}
    \label{Fig:RoomMesh}
\end{figure}

\begin{figure}
\centering
    \subfigure[]{\includegraphics[width=0.32\textwidth]{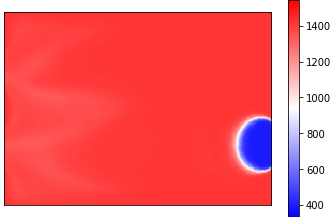}}
    \subfigure[]{\includegraphics[width=0.32\textwidth]{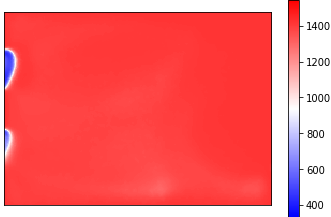}}
    \subfigure[]{\includegraphics[width=0.32\textwidth]{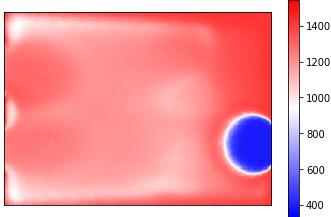}}
    \subfigure[]{\includegraphics[width=0.32\textwidth]{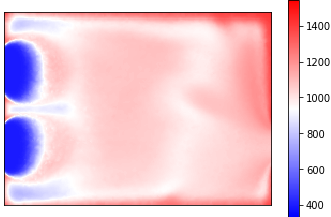}}
    \subfigure[]{\includegraphics[width=0.32\textwidth]{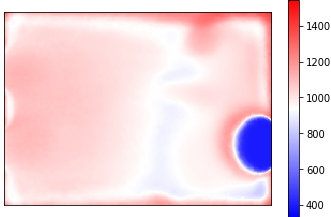}}
    \subfigure[]{\includegraphics[width=0.32\textwidth]{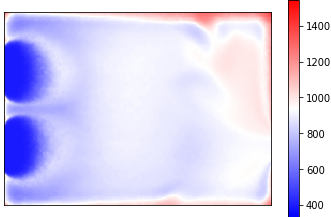}}
\caption{``Structured dataset" coloured by CO\textsubscript{2} concentration at timesteps (a) 509, (b) 643 (c) 1062, (d) 1359, (e) 1485 and (f) 1741. Blue colour means low CO\textsubscript{2} concentration, i.e. 400 ppm, and red colour indicates high value of concentration, i.e. 1420 ppm.}
\label{Fig:StructuredDS}
\end{figure}

\begin{figure}
\centering
    \subfigure[]{\includegraphics[width=0.32\textwidth]{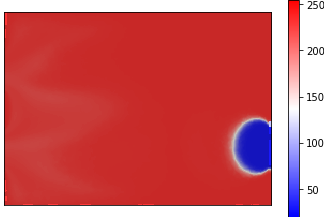}}
    \subfigure[]{\includegraphics[width=0.32\textwidth]{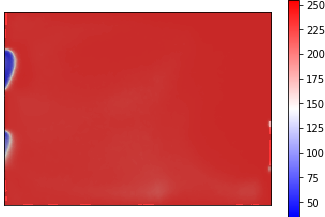}}
    \subfigure[]{\includegraphics[width=0.32\textwidth]{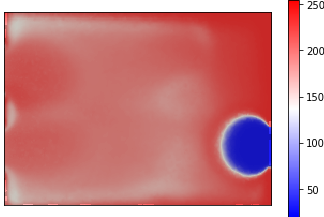}}
    \subfigure[]{\includegraphics[width=0.32\textwidth]{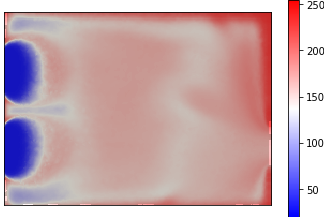}}
    \subfigure[]{\includegraphics[width=0.32\textwidth]{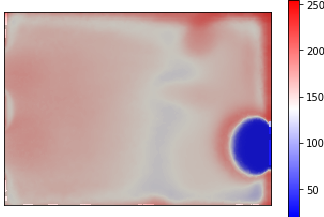}}
    \subfigure[]{\includegraphics[width=0.32\textwidth]{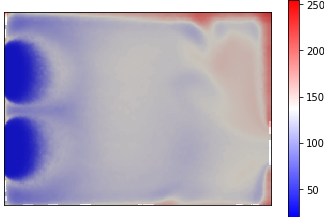}}
\caption{``RGB dataset" colored by RGB values scaled using the CO\textsubscript{2} concentration at timesteps (a) 509, (b) 643 (c) 1062, (d) 1359, (e) 1485 and (f) 1741. Blue colour means low CO\textsubscript{2} concentration, i.e. 400 ppm, and corresponds to RGB values equal to 0. Red colour indicates high value of concentration, i.e. 1420 ppm, and corresponds to RGB values equal to 255.}
\label{Fig:rgb_cut}
\end{figure}

The data generated by the CFD simulation are stored on an unstructured mesh and need to be converted into structured data in order to apply the LA model presented in Section~\ref{Sec:LA}. Indeed, convolutional kernels work on the assumption that adjacent states are equally spaced. As a first step, only the nodes located within the test room were selected to work with, thus excluding the rest of the domain to the working dataset as shown in Figure~\ref{Fig:RoomMesh}. As a second step, we choose to extract and work with the data from a 2D slice located at half height of the room: this location being a good compromise between the different heights of sensors used during the field experiment. Finally, two different pre-processing approaches were adopted:
\begin{itemize}
    \item \textbf{``Structured dataset"} Data from the unstructured 2D slice are projected on a structured grid. The ``Structured dataset" is generated by interpolating the CO\textsubscript{2} concentration values of the unstructured grid on a structured grid. The values stored in the final matrix corresponds to actual values of CO\textsubscript{2} concentration as shown in Figure~\ref{Fig:StructuredDS}.
    \item \textbf{``RGB dataset"} Data from the unstructured 2D slice are directly converted into a RGB image: a screenshot of the 2D slice coloured based on the CO\textsubscript{2} concentration values is created. The scalar bar of the RGB images is set based on the minimum and maximum CO\textsubscript{2} concentration, i.e. 400 ppm and 1420 ppm, respectively. This transformation allows to move from unstructured mesh to structured data since the RGB image is a 3D structured matrix of pixel values. The values stored in the final matrix corresponds to RGB values being between 0 and 255 as shown in Figure~\ref{Fig:rgb_cut}.
\end{itemize}

Both pre-processing approaches are performed for each timestep and the final size of the working matrix is a 180$\times$250 regular grid.

As a final step, all the data are normalised between 0 and 1. The ``RGB dataset" is divided by 255, while the ``Structured dataset" is normalised based on the minimum and maximum values of CO\textsubscript{2} concentration.

\subsection{Sensors data pre-processing}
\begin{figure}[H]
\centering
    \subfigure[]{\includegraphics[width=0.32\textwidth]{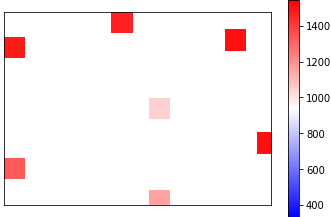}}
    \subfigure[]{\includegraphics[width=0.32\textwidth]{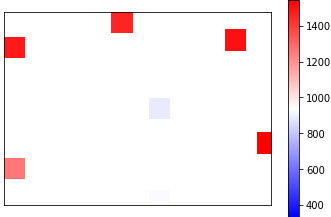}}
    \subfigure[]{\includegraphics[width=0.32\textwidth]{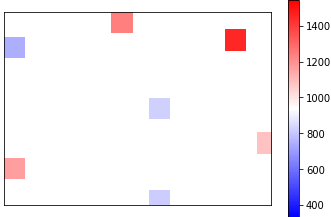}}
    \subfigure[]{\includegraphics[width=0.32\textwidth]{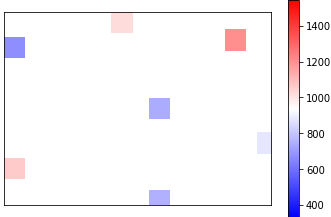}}
    \subfigure[]{\includegraphics[width=0.32\textwidth]{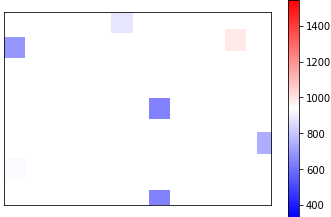}}
    \subfigure[]{\includegraphics[width=0.32\textwidth]{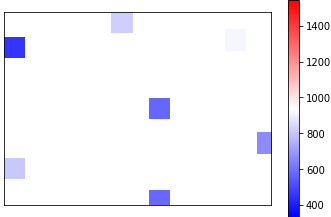}}
\caption{Rendering of the sensor values expanding to a radius of 15 cm coloured by CO\textsubscript{2} concentration (in ppm) at timesteps (a) 509, (b) 643, (c) 1062, (d) 1359, (e) 1485 and (f) 1741.}
\label{Fig:sensor_pos}
\end{figure}

\begin{figure}[H]
\centering
    \subfigure[]{\includegraphics[width=0.32\textwidth]{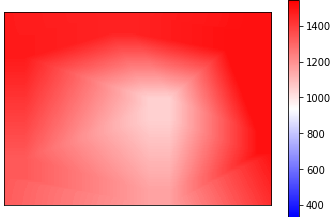}}
    \subfigure[]{\includegraphics[width=0.32\textwidth]{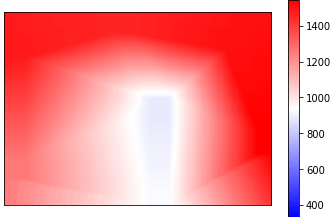}}
    \subfigure[]{\includegraphics[width=0.32\textwidth]{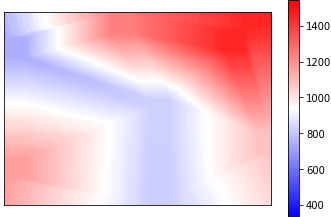}}
    \subfigure[]{\includegraphics[width=0.32\textwidth]{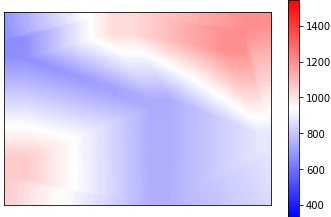}}
    \subfigure[]{\includegraphics[width=0.32\textwidth]{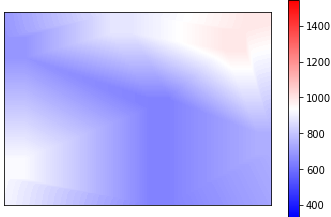}}
    \subfigure[]{\includegraphics[width=0.32\textwidth]{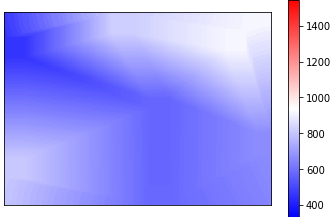}}
\caption{Rendering of the linear interpolation of the sensor values coloured by CO\textsubscript{2} concentration (in ppm) at timesteps (a) 509, (b) 643, (c) 1062, (d) 1359, (e) 1485 and (f) 1741.}
\label{Fig:sensor_prep}
\end{figure}

The location of the sensors during the field experiment are shown in Figure~\ref{Fig:ClarenceCentre} and are reported in our 180$\times$250 matrix at the corresponding timestep. Based on the sampling rate of the sensors, 10 CFD output were selected corresponding to time levels for which we have sensors data. As a first pre-processing step, considering that the area of influence of one sensor has a radius of about 15 cm, a zone of 10 pixels $\times$ 10 pixels centred on the sensor location is defined and the value given by the sensor at that location is assigned to this whole area. The rendering of this process is shown in Figure~\ref{Fig:sensor_pos}. The second step consists in interpolating linearly the values of the sensors to the entire 2D structured grid as shown in Figure~\ref{Fig:sensor_prep}. This pre-processing is performed to be consistent with both the ``Structured dataset" and the ``RGB dataset", i.e. is done in terms of CO\textsubscript{2} concentration and RGB values scaled using the CO\textsubscript{2} concentration, respectively. 
As a final step, all the data are normalised between 0 and 1 as for the CFD pre-processing.

\section{Results and Discussions}
In this section, the procedure to determine the optimal network architectures of both the AutoEncoder and the LSTM is first presented. Then the results of the novel Latent Assimilation model developed in this paper are discussed based on the assimilation of the sensors data in both latent and physical space.

\begin{figure}
    \centering
    \includegraphics[width=1\textwidth]{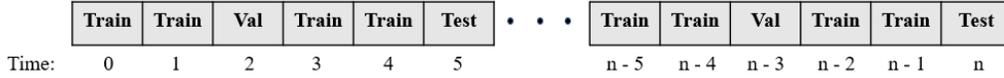}
    \caption{Training, Validation and Testing sets. Two consecutive timesteps are considered for the training. The jumped data are assigned to the validation and testing sets alternately.}
    \label{Fig:data_split}
\end{figure}

The data set is decomposed into training, validation and testing sets. In the CFD simulation, the flow field and the associated CO\textsubscript{2} concentration does not change much between consecutive steps. For this reason, we decide to divide the data in training, validation and testing sets making jumps. First, the CFD output at timesteps corresponding to sensors data are excluded and are assigned to the testing set. For the remaining data, two consecutive timesteps are considered for the training, then a jump is performed. The jumped data, i.e the ones not considered yet, are assigned to the validation and testing sets alternately. Considering a jump equal to 1, this process is summarised in Figure~\ref{Fig:data_split}.

\subsection{AutoEncoder network architecture}
The AutoEncoder implemented in this paper is a Convolutional AutoEncoder (CAE). Specifically, the encoder is composed by several convolutional layers followed by a flattened layer then a regular densely-connected layer which determine the shape of the latent space. In our CAE, the decoder architecture has almost the same structure than the encoder one: indeed, an additional convolutional layer is used in the decoder. Finding the optimal construction of the CAE architecture is divided in two steps:
\begin{enumerate}
    \item Finding the optimal numbers of layers
    \item Grid search: Finding the optimal hyperparameters
\end{enumerate}

For each CAE network architecture tested, 5-fold cross-validation are performed for which the data are shuffled to make the neural network independent from the order of the data. Both the training and validation sets are used for the cross-validation. The evaluation of the CAE network architecture is estimated based on the mean and the standard deviation of the Mean Squared Error between the CAE prediction and the CFD output, i.e Mean-MSE and Std-MSE, respectively; the mean and the standard deviation of the Mean Absolute Error between the CAE prediction and the CFD output, i.e Mean-MAE and Std-MAE; and the mean and the standard deviation of the CAE execution time, i.e. Mean-Time and Std-Time. A low MSE/MAE standard deviation reflects that the model is stable and does not depend on the data used to train and validate it, while a low MSE/MAE means that the prediction is close to the real input, i.e has a good accuracy.

\subsubsection{Optimal structure and number of layers}
The \textit{baseline} CAE network architecture is using the following fixed parameters:
\begin{itemize}
    \item \textbf{Convolutional layers parameters} 
    \begin{itemize}
        \item Number of filters: 32
        \item Activation function: Last decoder layer: sigmoid function to restrict the output in a range [0, 1] as the input; Rectified Linear Unit (ReLU) otherwise.
    \end{itemize}
    \item \textbf{Regular densely-connected layer} Latent Space: 7
    \item \textbf{Training configuration} 
    \begin{itemize}
        \item Losses and metrics optimiser: Adam, learning rate of $1.10^{-3}$. Adam optimisation is a stochastic gradient descent method based on adaptive estimation of first-order and second-order moments well suited for problems with large data.
        \item Number of epochs: 300
        \item Batch size: 32
    \end{itemize}
\end{itemize}

The following different CAE network architectures are tested using the ``Structured dataset" in order to find the optimal numbers and structure of layers:
\begin{enumerate}
    \item Encoder: 3 convolutional layers. Decoder: 4 convolutional layers. 3$\times$3 kernel size.
    \item Encoder: 4 convolutional layers. Decoder: 5 convolutional layers. 3$\times$3 kernel size.
    \item Encoder: 5 convolutional layers. Decoder: 6 convolutional layers. 3$\times$3 kernel size.
    \item Encoder: 4 convolutional layers. Decoder: 5 transpose convolutional layers. 3$\times$3 kernel size.
    \item Encoder: 4 convolutional layers. Decoder: 5 convolutional layers. 5$\times$5 kernel size.
    \item Encoder: 4 convolutional layers; 5$\times$5, 5$\times$5, 3$\times$3 and 3$\times$3 kernel sizes. Decoder: 5 convolutional layers; 3$\times$3, 3$\times$3, 5$\times$5, 5$\times$5 and 5$\times$5 kernel sizes.
\end{enumerate}

Results of the evaluation of the six CAE network architectures are reported in Table~\ref{Tab:conf_2} where the configuration number N are given by the list above.

\begin{table}[H]
    \centering
    \resizebox{\textwidth}{!}{
        \begin{tabular}{@{}ccccccc@{}}
            \toprule
            \textbf{N} & \textbf{Mean-MSE} & \textbf{Std-MSE} & \textbf{Mean-MAE} & \textbf{Std-MAE} & \textbf{Mean-Time} & \textbf{Std-Time} \\ \midrule
            1 & 1.324e-02 & 2.072e-02 & 5.573e-02 & 7.575e-02 & 775.665 & 5.606e+00 \\ \midrule
            \textbf{2} & \textbf{2.435e-04} & \textbf{3.851e-05} & \textbf{1.034e-02} & \textbf{8.776e-04} & \textbf{812.293} & \textbf{1.436e+00} \\ \midrule
            3 & 2.293e-02 & 2.763e-02 & 8.887e-02 & 9.818e-02 & 828.328 & 1.574e+00 \\ \midrule
            4 & 2.587e-04 & 4.114e-05 & 1.041e-02 & 9.384e-04 & 746.804 & 3.089e+00 \\ \midrule
            5 & 1.055e-02 & 2.099e-02 & 4.460e-02 & 7.922e-02 & 1222.251 & 6.627e+00 \\ \midrule
            6 & 1.058e-02 & 2.094e-02 & 4.650e-02 & 7.826e-02 & 1164.464 & 4.605e+00\\ \bottomrule
        \end{tabular}
    } \caption{Convolutional AutoEncoder performance evaluation for 6 network architectures. N denotes the configuration number as listed in the main text. Time is given in seconds. The bold row, i.e. configuration 2, is the configuration highlighting the best overall performance.} \label{Tab:conf_2}
\end{table}

\textbf{Number of layers} Comparing configurations 1, 2 and 3 for which only the number of convolutional layers is changing, configuration 2, i.e. 4 convolutional layers for the encoder and 5 convolutional layers for the decoder, is the one highlighting the best performance in terms of both Mean-MSE and Mean-MAE with MSE two order of magnitude lower than configurations 1 and 3. Moreover, configuration 2 is the most stable regarding the standard deviations, reflecting well that this CAE network architecture does not depend on the data used to train and validate it. In addition, the execution time of configuration 2 is relatively acceptable to answer real-time problems. Hence, in the following, the number of layers is taken as the same than configuration 2: 4 for the encoder and 5 for the decoder.

\textbf{Convolutional vs transpose convolutional layers in the decoder} The accuracy (Mean-MSE/Mean-MAE) and the stability (Std-MSE/Std-MAE) are slightly better, while the execution time is slightly longer, when using convolutional layers (config. 2) rather than transpose convolutional layers (config. 4) in the decoder. As no major improvements in terms of MSE/MAE is observed when switching from convolutional (config. 2) to transpose convolutional layers in the decoder (config. 4), convolutional layers are used for the decoder.

\textbf{Size of the kernel} Configurations 2, 5 and 6 have the same layer number and the size of the kernel is changed. Using a 5$\times$5 kernel size for all the layers (config. 5) or using a mix of 3$\times$3 and 5$\times$5 kernel sizes (config. 6) both increase the MSE/MAE by two order of magnitude compared to using a 3$\times$3 kernel size for all the layers (config. 2). In addition, the execution cost is considerably increased, about 50 \% less efficient, when the kernel size is larger as the complexity scales with $k^3$ where $k$ is the kernel size. Overall, 3$\times$3 is then used as the optimal kernel size in the following.

\subsubsection{Grid search for the optimal hyperparameters}\label{Sec:AE_Optimal}
The grid search is now performed in order to find the optimal hyperparameters for both the ``Structured dataset" and the ``RGB dataset". The CAE network architecture is using the following fixed parameters:
\begin{itemize}
    \item \textbf{Convolutional layers parameters} 
    \begin{itemize}
        \item Number of encoder/decoder convolutional layers: 4 and 5
        \item Kernel size: 3x3
    \end{itemize}
    \item \textbf{Regular densely-connected layer} Latent Space: 7
    \item \textbf{Training configuration} Adam optimiser: learning rate of $1.10^{-3}$.
\end{itemize}

The hyperparameters tested for the grid search are as follows: 
\begin{itemize}
    \item \textbf{Convolutional layers parameters} 
    \begin{itemize}
        \item Number of filters: 16, 32, 64
        \item Activation function: Rectified Linear Unit (ReLU), Exponential Linear Unit (ELU).
    \end{itemize}
    \item \textbf{Training configuration} 
    \begin{itemize}
        \item Number of epochs: 250, 300, 400
        \item Batch size: 16, 32, 64
    \end{itemize}
\end{itemize}

Table~\ref{Tab:AE_hyper} shows the optimal hyperparameters found for each input dataset, while the evaluation performance are reported in Table~\ref{Tab:AE_performance}. The optimal hyperparameters are the same for both dataset, i.e. 64 number of filters, a ReLU activation function and 400 epochs. Only the batch size differs: 32 for the ``Structured dataset" and 16 for the ``RGB dataset". The fact that ``RGB dataset" needs less batch size than the ``Structured dataset" can be potentially attributed to the fact that the former has 3 channels (R, G and B colours), so requiring less batch size. The results show that both datasets have very similar accuracy and stability: low MSE and low standard deviation, of the order of $10^{-5}$, meaning that the CAE is not dependent on the set of input chosen to train it. Using ``RGB dataset" highlights better performance, with Mean-MSE 57 \% lower than when using ``Structured dataset". However, using the ``RGB dataset",  more time is needed to train the CAE because an element of this dataset is composed by three channels, i.e. the R, G and B colour values.

\begin{table}[H]
    \centering
    \resizebox{0.7\textwidth}{!}{
        \begin{tabular}{@{}ccccc@{}}
            \toprule
            \textbf{Dataset} & \textbf{Filters} & \textbf{Activation} & \textbf{Epochs} & \textbf{Batch size} \\ \midrule
            \textbf{Structured} & 64 & ReLU & 400 & 32 \\ \midrule
            \textbf{RGB} & 64 & ReLU & 400 & 16 \\ \bottomrule
        \end{tabular}
    }\caption{Optimal Convolutional AutoEncoder hyperparameters determined by a grid search when using ``Structured dataset" or ``RGB dataset" as input.}\label{Tab:AE_hyper}
\end{table}

\begin{table}[H]
    \centering
    \resizebox{\textwidth}{!}{%
        \begin{tabular}{@{}lcccccc@{}}
            \toprule
            \textbf{Dataset} & \textbf{Mean-MSE} & \textbf{Std-MSE} & \textbf{Mean-MAE} & \textbf{Std-MAE} & \textbf{Mean-Time} & \textbf{Std-Time} \\ \midrule
            \textbf{Structured} & 8.509e-05 & 1.577e-05 & 6.182e-03 & 8.714e-04 & 1887.612 & 6.845e+00 \\ \midrule
            \textbf{RGB} & 3.670e-05 & 9.261e-06 & 3.601e-03 & 4.333e-04 & 2490.920 & 1.825e+01 \\ \bottomrule
        \end{tabular}%
    } \caption{Convolutional AutoEncoder performance using ``Structured dataset" or ``RGB dataset" as input and the optimal hyperparameters found with the grid search (Table~\ref{Tab:AE_hyper}. Time is given in seconds.} \label{Tab:AE_performance}
\end{table}

\subsection{LSTM network architecture}
\begin{figure}[H]
    \centering
    \includegraphics[width=0.7\textwidth]{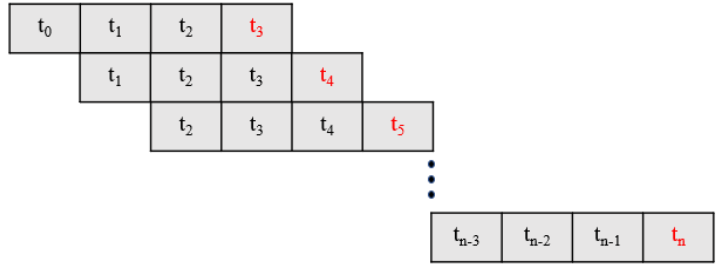}
    \caption{Data splitting for LSTM model. Timesteps in black are the ones used to make the prediction and the red ones are the timesteps predicted.}
    \label{Fig:lstm}
\end{figure}

In this model, the training set is used for the fitting step and the validation set for the validation step. An extra splitting of the training set is performed: the data are split in small sequences such that one timestep is predicted and 3 timesteps are used as ``look back" values as shown in Figure~\ref{Fig:lstm}. All data are encoded with the AutoEncoder: each input sample of the LSTM is then a vector of 7 scalars. Finding the optimal construction of the LSTM architecture is divided in two steps:
\begin{enumerate}
    \item Finding the optimal numbers of layers
    \item Grid search: Finding the optimal hyperparameters
\end{enumerate}

For each LSTM network architecture tested, the fitting and the evaluation of the model is repeating 5 times. As for the CAE network architecture, the LSTM network architecture performance is evaluated based on the Mean-MSE, Std-MSE, Mean-MAE, Std-MAE, Mean-Time and Std-Time.

\subsubsection{Optimal number of layers}
The \textit{baseline} LSTM network architecture, tested with the ``Structured dataset", is using the following fixed parameters:
\begin{itemize}
    \item \textbf{LSTM layers parameters}
    \begin{itemize}
        \item Number of neurons: 30, i.e the dimensionality of the output space
        \item Activation function: ReLU
        \item Number of steps: 3
    \end{itemize}
    \item \textbf{Regular densely-connected layer} Latent Space: 7
    \item \textbf{Training configuration}
    \begin{itemize}
        \item Optimiser: Adam with a learning rate of $1.10^{-3}$
        \item Number of epochs: 300
        \item Batch size: 32
    \end{itemize}
\end{itemize}

LSTMs are stacked from 1 to 5 times in order to see if the model gains in accuracy, stability and efficiency: the results are shown in Table~\ref{Tab:lstm_lay}. The single layer LSTM is the one highlighting the best accuracy with the lowest Mean-MSE and Mean-MAE values. Indeed, the input of the LSTM consists of a 7$\times$1 vector and adding more LSTM layer introduces overfitting bias. In addition, the standard deviation, reflecting the stability, of the single layer LSTM are about one order of magnitude lower than the other tested LSTM. Finally, as expected, the single layer LSTM is also the most efficient in term of computation cost.

\begin{table}[H]
    \centering
    \resizebox{\textwidth}{!}{%
        \begin{tabular}{@{}ccccccc@{}}
            \toprule
            \textbf{N} & \textbf{Mean-MSE} & \textbf{Std-MSE} & \textbf{Mean-MAE} & \textbf{Std-MAE} & \textbf{Mean-Time} & \textbf{Std-Time} \\ \midrule
            \textbf{1} & \textbf{1.634e-02} & \textbf{2.510e-03} & \textbf{7.822e-02} & \textbf{8.553e-03} & \textbf{230.355} & \textbf{1.050e+00} \\ \midrule
            2 & 2.822e-02 & 7.244e-03 & 1.060e-01 & 1.679e-02 & 360.877 & 6.618e-01 \\ \midrule
            3 & 4.619e-02 & 1.942e-02 & 1.381e-01 & 3.034e-02 & 494.254 & 2.258e+00 \\ \midrule
            4 & 5.020e-02 & 1.675e-02 & 1.414e-01 & 2.778e-02 & 658.039 & 2.632e+00 \\ \midrule
            5 & 4.742e-02 & 1.183e-02 & 1.459e-01 & 2.075e-02 & 806.001 & 5.921e+00 \\ \bottomrule
        \end{tabular}%
    } \caption{LSTM performance evaluation for 5 network architectures. N denotes the number of stacked LSTMs. Time is given in seconds. The bold row, i.e. 1 LSTM, is the configuration highlighting the best overall performance.} \label{Tab:lstm_lay}
\end{table}

\subsubsection{Grid Search}
The grid search is now performed in order to find the optimal hyperparameters for both the ``Structured dataset" and the ``RGB dataset". The LSTM network architecture is using the following fixed parameters:
\begin{itemize}
    \item \textbf{LSTM layers parameters} Number of layers: 1
    \item \textbf{Regular densely-connected layer} Latent Space: 7
    \item \textbf{Training configuration} Optimiser: Adam, learning rate of $1.10^{-3}$
\end{itemize}

The hyperparameters tested for the grid search are as follows: 
\begin{itemize}
    \item \textbf{LSTM layers parameters}
    \begin{itemize}
        \item Number of neurons: 30, 50, 70
        \item Activation function: ReLU, ELU
        \item Number of steps: 3, 5, 7
    \end{itemize}
    \item \textbf{Training configuration}
    \begin{itemize}
        \item Number of epochs: 200, 300, 400
        \item Batch size: 16, 32, 64
    \end{itemize}
\end{itemize}

Table~\ref{Tab:LSTM_hyperparameters} shows the optimal hyperparameters found for each input dataset, while the evaluation performance are reported in Table~\ref{Tab:LSTM_performance}. Exponential Linear Unit (ELU) appears to be the optimal activation function, with an epochs of 400 and a batch size of 16  for both the input dataset. The results show that the ``RGB dataset" needs more neurons and more back observations than the ``Structured dataset". From Table~\ref{Tab:LSTM_performance}, it can be seen that the LSTM with ``RGB dataset" as input has better accuracy and takes also less time than when using ``Structured dataset" input. Indeed, the accuracy is about 45 \% higher when using input ``RGB dataset", while the execution time is reduced by approximately 27 \%.

\begin{table}[H]
    \centering
    \resizebox{0.8\textwidth}{!}{%
        \begin{tabular}{@{}ccclcc@{}}
            \toprule
            \textbf{Dataset} & \textbf{Neurons} & \textbf{Activation} & \textbf{Steps} & \textbf{Epochs} & \textbf{Batch size} \\ \midrule
            \textbf{Structured} & 30 & Elu & 3 & 400 & 16 \\ \midrule
            \textbf{RGB} & 50 & Elu & 7 & 400 & 16 \\ \bottomrule
        \end{tabular}%
    } \caption{Optimal LSTM hyperparameters determined by a grid search when using ``Structured dataset" or ``RGB dataset" as input.} \label{Tab:LSTM_hyperparameters}
\end{table}

\begin{table}[H]
    \centering
    \resizebox{\textwidth}{!}{%
        \begin{tabular}{@{}lcccccc@{}}
            \toprule
            \textbf{DataSet} & \textbf{Mean-MSE} & \textbf{Std-MSE} & \textbf{Mean-MAE} & \textbf{Std-MAE} & \textbf{Mean-Time} & \textbf{Std-Time} \\ \midrule
            \textbf{Structured} & 1.233e-02 & 1.398e-03 & 6.743e-02 & 3.428e-03 & 949.328 & 7.508e+00 \\ \midrule
            \textbf{RGB} & 6.712e-03 & 6.704e-04 & 4.661e-02 & 3.004e-03 & 690.594 & 2.743e+00 \\ \bottomrule
        \end{tabular}%
    } \caption{LSTM performance using ``Structured dataset" or ``RGB dataset" as input and the optimal hyperparameters found with the grid search (Table~\ref{Tab:LSTM_hyperparameters}). Time given in seconds.} \label{Tab:LSTM_performance}
\end{table}

\subsection{Latent Assimilation model}
In this section, results of our novel Latent Assimilation (LA) model are presented: the assimilation takes place in the latent space. The Testing set is considered and both dataset, i.e. ``Structured dataset" and ``RGB dataset", are encoded using the AutoEncoders with optimal network architecture as presented in Section~\ref{Sec:AE_Optimal}. The predictions are performed through the LSTM and are updated using the corresponding observations through Optimal Interpolated Kalman Filter (KF). 

In the KF, the error covariance matrix $\hat{Q}$ is computed as $\hat{Q}=VV^T$, where $V$ is as defined in equation~\eqref{eq:covarianProcedure}. 
Since both predictions of the model and observations are values of CO\textsubscript{2} or pixels, i.e. the observations do not have to be transformed, the operator $\hat{H}$ is an identity matrix. We studied how KF improves the accuracy of the prediction by testing different forms of the observation error covariance matrix $\hat{R}$: computed using equation~\eqref{eq:covarianProcedure} or, fixed as $\hat{R}=0.01I$, $0.001I$, $0.0001I$ where $I\in \mathbb{R}^{p\times p}$ denotes the identity matrix. This last assumption is usually made to give higher fidelity and trust to the observations~\cite{asch2016data}.

\subsubsection{Latent space}
The MSE between the background data and the observed data in the latent space for the ``Structured dataset" and the ``RGB dataset", without performing data assimilation, are $7.220\times10^{-1}$ and $5.447\times10^{-1}$, respectively. Table~\ref{Tab:LA_performance_LatentSpace} shows values of MSE in the latent space between the assimilated data $h^a_t$ and the observed data as well as the execution time of the assimilation for both input dataset. As expected, we can observe an improvement in the execution time of the assimilation in assuming $\hat{R}$ as a diagonal matrix instead of a full matrix. In addition, the assimilation increases the accuracy of the model, whatever the input dataset used, with MSE values about 2.2 times lower compared to without assimilation, highlighting that our novel Latent Assimilation model is behaving as expected. Using $\hat{R}$ as an identity matrix of the form $0.0001I$ allows to improve the accuracy by up to 4 order of magnitude.

\begin{table}[H]
    \centering
    \resizebox{0.9\textwidth}{!}{%
        \begin{tabular}{cccccccc}
            \cline{3-6}
            \multicolumn{2}{c}{} & \multicolumn{4}{c}{${\bf \hat{R}} $} \\ \cline{3-6}
            \multicolumn{2}{c}{} & \textbf{cov-matrix eq.~\eqref{eq:covarianProcedure}} & \textbf{0.01I} & \textbf{0.001I} & \textbf{0.0001I} \\ \midrule
            \multirow{2}{*}{\textbf{MSE}} & \textbf{Structured} & 3.215e-01 & 1.250e-02 & 1.787e-03 & 3.722e-05\\ 
             &\textbf{RGB} & 2.444e-01 & 1.002e-02 & 4.409e-03 & 1.640e-03 \\ \midrule
            \multirow{2}{*}{\textbf{Time}} &\textbf{Structured} & 1.541e-03 & 5.410e-04 & 4.854e-04 & 4.823e-04 \\
            &\textbf{RGB} & 2.145e-03 & 5.388e-04 & 4.847e-04 & 4.807e-04 \\ \bottomrule
        \end{tabular}
    } \caption{MSE values in the \textbf{latent space} and execution time of the assimilation in seconds of the Latent Assimilation model for different form of the observations error covariance matrix $\hat{R}$ in the latent space when using the ``Structured dataset" or the ``RGB dataset" as input.} \label{Tab:LA_performance_LatentSpace}
\end{table}

\subsubsection{Physical space}
\begin{figure}[H]
    \centering
    \subfigure[][Predicted state t=509]{\includegraphics[width=0.32\textwidth]{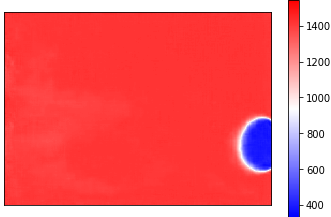}}
    \subfigure[][Observation t=509]{\includegraphics[width=0.32\textwidth]{img/sensor_0.png}}
    \subfigure[][Updated state t=509]{\includegraphics[width=0.32\textwidth]{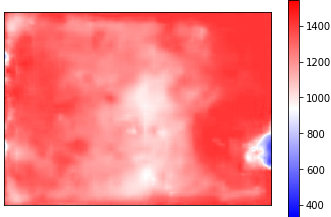}}
    \subfigure[][Predicted state t=1062]{\includegraphics[width=0.32\textwidth]{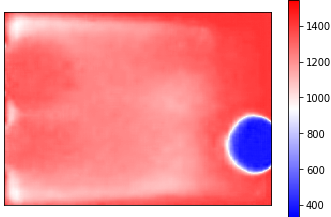}}
    \subfigure[][Observation t=1062]{\includegraphics[width=0.32\textwidth]{img/sensor_2.png}}
    \subfigure[][Updated state t=1062]{\includegraphics[width=0.32\textwidth]{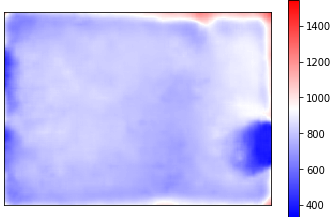}} 
    \subfigure[][Predicted state t=1485]{\includegraphics[width=0.32\textwidth]{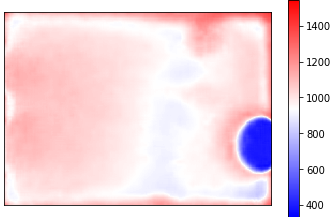}}
    \subfigure[][Observation t=1485]{\includegraphics[width=0.32\textwidth]{img/sensor_4.png}}
    \subfigure[][Updated state t=1485]{\includegraphics[width=0.32\textwidth]{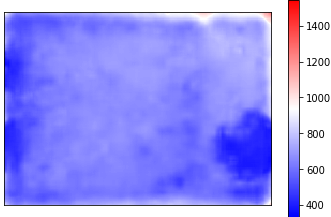}} 
    \caption{Assimilation results colored by CO\textsubscript{2} concentration (in ppm) at timesteps 509, 1062 and 1485. (a) LSTM predicted state, (b) interpolated observation and (c) updated state obtained using the latent assimilation. Blue colour means low CO\textsubscript{2} concentration, i.e. 400 ppm, and red colour indicates high value of concentration, i.e. 1420 ppm.}
    \label{Fig:assimilation_509}
\end{figure}

After having performed the DA in the latent space, the results $h^a_t$ are reported in the physical space through the decoder which gives $x^a_t$. Figure~\ref{Fig:assimilation_509} shows in the physical space the results of the assimilation for the timesteps 509, 1062 and 1485 using our novel LA model. 

The MSE in the physical space using our LA model is then compared with the one using a standard Data Assimilation (sDA) procedure. sDA is performed in the physical space using a Kalman Filter approach (equations~\eqref{Eq: StartKalman}-~\eqref{eq:kalmananalysis}), where $R \in \mathbb{R}^{n\times n}$ is defined in the physical space. Table~\ref{Tab:LA_performance_PhysicalSpace} shows values of MSE in the physical space between the assimilated data $x^a_t$ and the observed data as well as the execution time of the assimilation for our LA model and the standard methodology (sDA) when using the ``Structured dataset" as input. The MSE between the background data and the observed data in the physical space, without performing data assimilation, is $6.491\times10^{-2}$. Both LA and sDA improve the accuracy of the forecasting as shown in Table~\ref{Tab:LA_performance_PhysicalSpace}: however it can be observed that the LA model gives 35 \% more accuracy than a sDA model. In addition, LA performs better in terms of execution time with respect to a sDA: indeed sDA works directly with big matrices making it slower by six order of magnitude. 

\begin{table}[H]
    \centering
    \resizebox{0.9\textwidth}{!}{%
        \begin{tabular}{cccccccc}
            \cline{3-6}
            \multicolumn{2}{c}{} & \multicolumn{4}{c}{${\bf \hat{R}} $} \\ \cline{3-6}
            \multicolumn{2}{c}{} & \textbf{cov-matrix eq.~\eqref{eq:covarianProcedure}} & \textbf{0.01I} & \textbf{0.001I} & \textbf{0.0001I}  \\ \midrule
            \multirow{2}{*}{\textbf{MSE}} & \textbf{LA} & 3.356e-02 & 6.933e-04 & 1.211e-04 & 2.691e-06 \\ 
             &\textbf{sDA} & 5.179e-02 & 6.928e-03 & 6.928e-03 & 6.997e-03 \\ \midrule
            \multirow{2}{*}{\textbf{Time}} &\textbf{LA} & 1.541e-03 & 5.410e-04 & 4.854e-04 & 4.823e-04 \\
            &\textbf{sDA} & 2.231e+03 & 2.148e+03 & 2.186e+03 & 2.159e+03 \\ \bottomrule
        \end{tabular}
    } \caption{MSE values of $x^a_t$ using our novel Latent Assimilation (LA) model or using a standard Data Assimilation (sDA) procedure for different form of the observations error covariance matrix $\hat{R}$ when using the ``Structured dataset" as input. MSE are computed in the \textbf{physical space}. Execution time of the assimilation in seconds.} \label{Tab:LA_performance_PhysicalSpace}
\end{table}

\subsubsection{Size of the latent space}
In this section, the impact of increasing the size of the latent space is discussed. Results are presented for the ``Structured dataset" only. Table~\ref{Tab:LS_latent} and Table~\ref{Tab:LS_physical} give the MSE values of the Latent Assimilation model with different latent space sizes, from 1000 to 20000, computed in the latent and physical space, respectively. The column ``No DA" reports the MSE values without the assimilation. Table~\ref{Tab:LS_latent_time} reports the execution time of the assimilation.

 Defining $\hat{R}$ as an identity matrix always highlights better accuracy whatever the latent space size. Increasing the latent space size tends to decrease the MSE, i.e. gain in accuracy, in both the latent and the physical space. Overall, a latent space size equal to 18000 seems optimal for this problem, whatever the form of the observations error covariance matrix $\hat{R}$, which represents about 40 \% of the original data. However, the execution time of the assimilation can be up to 5 order of magnitude higher when using the optimal latent space size compared to a lower size. Finding the optimal parameters of our LA depends the expectancy of the user as a balance needs to be taken between accuracy and efficiency. Regarding the small accuracy gain while increasing the latent space size, it is recommended to work with the smallest latent space size as possible in order to benefit of the best efficiency while still keeping a high accuracy.

\begin{table}[H]
    \centering
    \resizebox{0.8\textwidth}{!}{%
        \begin{tabular}{@{}ccccccc@{}}
            \toprule
            \textbf{Latent Space} & \textbf{No DA} & \textbf{0.01\ I} & \textbf{0.001\ I} & \textbf{0.0001 \ I} \\ \midrule
            1000 & 4.363e-03 & 5.141e-04 & 4.980e-04 & 4.961e-04\\ \midrule
            3000 & 6.472e-04 & 9.933e-05 & 9.566e-05 & 9.501e-05\\ \midrule
            5000 & 6.012e-04 & 1.009e-04 & 9.184e-05 & 9.076e-05\\ \midrule
            7000 & 7.295e-04 & 1.179e-04 & 1.146e-04 & 1.142e-04 \\ \midrule
            12000 & 1.727e-04 & 3.651e-05 & 3.540e-05 & 3.531e-05\\ \midrule
            15000 & \textbf{1.660e-04} & 3.529e-05 & 3.463e-05 & 3.457e-05\\ \midrule
            18000 & 2.941e-04 & \textbf{3.141e-05} & \textbf{3.086e-05} & \textbf{3.085e-05} \\ \midrule
            20000 & 3.465e-04 & 6.506e-05 & 6.177e-05 & 6.073e-05\\ \bottomrule
        \end{tabular}
    } \caption{MSE values of LA model in the \textbf{latent space} with different latent space sizes. The column Latent Space indicates the size of the latent space, the column ``No DA" indicates the MSE of the model without the assimilation and the other columns indicate the MSE of the LA model with different observation error covariance matrix $\hat{R}$.} \label{Tab:LS_latent}
\end{table}

\begin{table}[H]
    \centering
    \resizebox{0.8\textwidth}{!}{%
        \begin{tabular}{@{}ccccccc@{}}
            \toprule
            \textbf{Latent Space} & \textbf{No DA} & \textbf{0.01\ I} & \textbf{0.001\ I} & \textbf{0.0001 \ I} \\ \midrule
            1000 & 3.532e-02 & 1.347e-03 & 1.278e-03 & 1.278e-03\\ \midrule
            3000 & 3.261e-02 & 1.828e-03 & 1.694e-03 & 1.653e-03\\ \midrule
            5000 & 2.822e-02 & 1.975e-03 & 1.689e-03 & 1.667e-03\\ \midrule
            7000 & 3.352e-02 & 1.248e-03 & 1.171e-03 & 1.155e-03\\ \midrule
            12000 & 2.479e-02 & 1.248e-03 & 1.171e-03 & 1.155e-03\\ \midrule
            15000 & \textbf{1.734e-02} & 1.325e-03 & 1.262e-03 & 1.253e-03\\ \midrule
            18000 & 3.703e-02 & \textbf{1.080e-03} & \textbf{9.848e-04} & \textbf{9.743e-04} \\ \midrule
            20000 & 2.514e-02 & 1.621e-03 & 1.424e-03 & 1.365e-03\\ \bottomrule
        \end{tabular}
    } \caption{MSE values of LA model in the \textbf{physical space} with different latent space sizes. The column Latent Space indicates the size of the latent space, the column ``No DA" indicates the MSE of the model without the assimilation and the other columns indicate the MSE of the LA model with different observation error covariance matrix $\hat{R}$.} \label{Tab:LS_physical}
\end{table}

\begin{table}[H]
    \centering
    \resizebox{0.7\textwidth}{!}{%
        \begin{tabular}{@{}cccccc@{}}
            \toprule
            \textbf{Latent Space} & \textbf{0.01\ I} & \textbf{0.001\ I} & \textbf{0.0001 \ I} \\ \midrule
            1000 & 5.637e-01 & 5.525e-01 & 5.637e-01 \\ \midrule
            3000 & 2.272e+00 & 2.260e+00 & 2.147e+00\\ \midrule
            5000 & 5.197e+00 & 5.301e+00 & 5.358e+00\\ \midrule
            7000 & 1.104e+01 & 1.125e+01 & 1.136e+01\\ \midrule
            12000 & 4.289e+01 & 4.353e+01 & 4.375e+01\\ \midrule
            15000 & 7.960e+01 & 8.072e+01 & 8.180e+01\\ \midrule
            18000 & 1.432e+02 & 1.441e+02 & 1.443e+02\\ \midrule
            20000 & 2.096e+02 & 2.124e+02 & 1.997e+02\\ \bottomrule
        \end{tabular}
    } \caption{Execution time in seconds of the assimilation in our LA model with different latent space sizes. The column Latent Space indicates the size of the latent space and the other columns indicate the execution time of the assimilation with different observation error covariance matrix $\hat{R}$.} \label{Tab:LS_latent_time}
\end{table}

\section{Conclusion and Future Work}\label{Sec: CFW}
In this paper, we proposed a new methodology called Latent Assimilation (LA) to efficiently and accurately perform Data Assimilation (DA). LA consists in performing the Optimal Kalman Filter in the latent space obtained by a Convolutional AutoEncoder with non-linear encoder functions and non-linear decoder functions. In the latent space, the dynamic system is represented by a surrogate model built by an LSTM network to train a function that emulates the dynamic system in the latent space. The data from the dynamic model and the real data coming from the sensors are both processed through the AutoEncoder. 

We applied the methodology to a real test case and we have shown that the LA performs better than a standard DA in terms of both accuracy and efficiency. The data of the real test case was time-series data representing the airflow within a naturally ventilated office room. The data was provided by CFD on an unstructured mesh and we pre-processed these data to extract two different structured datasets: one composed of 2D matrices of CO\textsubscript{2} concentration (``Structured dataset") and the other one composed of RGB images colored by CO\textsubscript{2} concentration (``RGB dataset"). We pre-processed also the data coming from sensors in the same manner.

We tried different AutoEncoder configurations and we performed a grid search for both input datasets in order to determine the optimal configurations. The same was done for the LSTM: it is the surrogate model. We performed the assimilation in the latent space using the Latent Assimilation model for both datasets as input. We tested also the standard data assimilation in the physical space and we have shown that LA performs better in terms of both efficiency and accuracy.

In conclusion, we have successfully proposed and developed a novel model able to assimilate data in the latent space, thus answering the needs of accuracy, stability and efficiency required by real-time systems. This methodology can be used for example to predict in real-time the load of virus, such as the SARS-COV-2, in indoor spaces by linking it to the concentration of CO\textsubscript{2}~\cite{peng2020exhaled}.

There are different improvements that could be applied to the model to be used with more challenging applications:
\begin{itemize}
    \item Develop an implementation of LA to emulate a variational DA~\cite{asch2016data} which is often applied to big data problems. In particular, we will focus on a 4D Variational (4DVar) method. 4DVar is a computational expensive method as it is developed to assimilate several observations (distributed in time) for each timestep of the forecasting model. We will develop an extended version of LA able to assimilate set of distributed observations for each timestep and, then, able to perform a 4DVar;
    \item Add a third dimension, i.e. test the methodology on a 3D space using a 3D Convolutional Autoencoder. Instead of cutting a slice, the 3D Convolutional Autoencoder will work on the complete room space without losing information;
    \item Recent research studies has started in the direction of working directly with unstructured meshes. It will be challenging developing Latent Assimilation with an Encoder-Decoder which works directly on a 3D adaptive and unstructured mesh;
    \item Instead of using only indoor data, the methodology could be applied considering the exchange with the outdoor environment or tested in different applications, i.e ocean.
\end{itemize}

\section*{Acknowledgments}
This work is supported by the EPSRC Grand Challenge grant Managing Air for Green Inner Cities (MAGIC) EP/N010221/1 and the EP/T003189/1 Health assessment across biological length scales for personal pollution exposure and its mitigation (INHALE).






\bibliographystyle{elsarticle-num-names}
\bibliography{elsarticle-num-names.bib}







\end{document}